\documentclass{article}

\usepackage[preprint]{neurips_2024}

\usepackage[utf8]{inputenc} 
\usepackage[T1]{fontenc}    
\usepackage{hyperref}       
\usepackage{url}            
\usepackage{booktabs}       
\usepackage{amsfonts}       
\usepackage{nicefrac}       
\usepackage{microtype}      
\usepackage[table]{xcolor}
\usepackage{minted}
\usepackage[breakable]{tcolorbox}
\tcbuselibrary{listings, skins}

\usepackage{wrapfig}
\usepackage{subcaption}
\usepackage{caption}
\usepackage{balance}
\usepackage{url}
\usepackage{graphicx}
\usepackage{amsmath}
\usepackage{amsthm}
\usepackage{booktabs}
\usepackage{algorithm}
\usepackage{algorithmic}
\usepackage{color}
\usepackage{xcolor}
\usepackage{xspace}
\usepackage[misc]{ifsym}
\newcommand{\model}{CoSineVerifier\xspace}

\usepackage[ruled, vlined, nofillcomment, linesnumbered, algo2e]{algorithm2e}

\usepackage{booktabs}
\usepackage{multirow}
\usepackage{balance}
\usepackage{enumitem}
\usepackage{stfloats}
\usepackage{diagbox}
\usepackage{fancyhdr}
\definecolor{result_color}{RGB}{250,250,210}
\usepackage{bm}
\usepackage[table]{xcolor}

\newcommand{\ignore}[1]{}

\title{CoSineVerifier: Tool-Augmented Answer Verification for Computation-Oriented Scientific Questions}

\author{
  \parbox{\linewidth}{\centering
    Ruixiang Feng\textsuperscript{2}\thanks{Equal contribution.}\hspace{4pt}\thanks{Work done during internship at Nanbeige Lab.},\;
    Zhenwei An\textsuperscript{1}\footnotemark[1],\;
    Yuntao Wen\textsuperscript{2},\;
    Ran Le\textsuperscript{1},\;
    Yiming Jia\textsuperscript{1},\; 
    Chen Yang\textsuperscript{1},\;\\
    Zongchao Chen\textsuperscript{1},\;
    Lisi Chen\textsuperscript{2},\;
    Shen Gao\textsuperscript{2}\thanks{Corresponding author.},\;
    Shuo Shang\textsuperscript{2}\footnotemark[3],\;
    Yang Song\textsuperscript{1}\footnotemark[3],\;
    Tao Zhang\textsuperscript{1}
    \vspace{0.5em} \\
    \normalfont{\textsuperscript{1}Nanbeige Lab, BOSS Zhipin\; \\
    \textsuperscript{2}University of Electronic Science and Technology of China, Chengdu, China\;
    \\
    {\{songyang,leran,anzhenwei\}@kanzhun.com},  {\{shengao, chenlisi\}@uestc.edu.cn}, \\ {\{fengrx02, jedi.shang\}@gmail.com}
  }}
}

\begin{document}

\noindent\includegraphics[height=0.8cm]{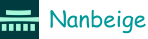}

\maketitle

\begin{abstract}
Answer verification methods are widely employed in language model training pipelines spanning data curation, evaluation, and reinforcement learning with verifiable rewards (RLVR). 
While prior work focus on developing unified verifiers applicable across multiple reasoning scenarios, significant challenges remain in computation-oriented scientific domains, such as algebraic equivalence checking and physical constant substitution.
In this paper, we introduce \model, a tool-augmented verifier that leverages external executors to perform precise computations and symbolic simplifications. 
\model enables robust verification that goes beyond simple semantic matching. 
We propose a novel two-stage pipeline, which begin with cold-start fine-tuning and followed by multi-turn reinforcement learning with tool integration.
Extensive experiments conducted on STEM subjects, general QA, and long-form reasoning tasks demonstrates strong generalization of \model.
The results shows that the \model achieves state-of-the-art performance on VerifyBench-Hard and SCI-Bench. 
And we also employ our \model in RLVR as a reward model, the results show that it consistently outperforms both rubric-based and model-based verifiers on AIME'24 and AIME'25, demonstrating strong potential to enhance reasoning capabilities of LLM.
Our model is released at \hyperlink{https://huggingface.co/Nanbeige/CoSineVerifier-Tool-4B}{https://huggingface.co/Nanbeige/CoSineVerifier-Tool-4B}.
\end{abstract}

\section{Introduction}\label{sec:intro}

The evolution of large language models (LLMs) training—from relying solely on human-annotated data to enabling models to self-evolve—has imposed increasingly stringent demands on how we evaluate the quality of LLM outputs. 
In this context, answer verification, which compares a model output against a reference, has become a critical component throughout data curation, evaluation, and model training pipelines.
In the phase of data curation, answer verification methods are extensively employed to guarantee the consistency, fidelity, and accuracy of synthetic corpora by many works~\cite{he2025deepmath,yuan2025naturalreasoning, toshniwal2024openmathinstruct,albalak2025big}.
At the evaluation stage, these methods provide a robust alternative to assess correctness beyond exact match (EM), enabling semantic-level comparison and format-robust scoring~\cite{ma2025s2r,li2024crowdsourced,paech2023eqbench,wu2025writingbench}.
For model training, Reinforcement Learning with Verifiable Rewards (RLVR) relies on verifiers to produce reward signals, imposing stringent requirements on verifier accuracy, calibration, and latency/throughput~\cite{he2025skywork, guha2025openthoughts, guo2025deepseek}. 

\begin{figure}[t]
        \centering
	\includegraphics[width=0.6\linewidth]  {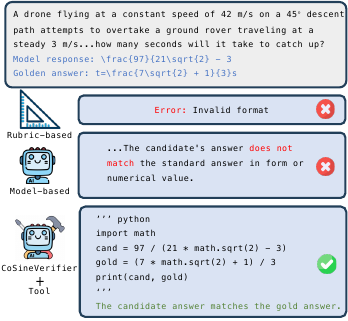}
        \caption{Comparison between \model and existing methods, which demonstrates our \model can enable accurate judgment on calculation-intensive scientific scenarios}
 \label{fig:introduction}
 \vspace{-13pt}
\end{figure}

Existing verification methods primarily rely on direct semantic comparisons between model outputs and reference answers.
Rubric-based verifiers~\cite{Kydlicek_Math-Verify_Math_Verification}, which depend on structured outputs (e.g., Math-Verify evaluating equivalence in Markdown or step-by-step formats), can fail even on simple computation-heavy cases such as comparing \textit{2*2} and \textit{4}.
Alternatively, many studies~\cite{chen2025xverify, liu2025compassverifier, chen2025judgelrm} employ LLM-based verifiers that assess answers through chain-of-thought (CoT) reasoning or implicit reasoning encoded in model parameters.
However, these verifiers are still prone to hallucination and inconsistencies in numerical or symbolic reasoning.
For instance, when comparing matrices with hundreds of entries or substituting physical constants with specific values, LLM-based verifiers often produce incorrect judgments.
The lack of robust computation skills prevents reasoning models from receiving accurate rewards in RLVR tasks within complex scientific domains, thereby limiting their overall potential.

We present \model, a tool-augmented LLM-based verifier designed for checking answers to computation-heavy scientific questions.
As shown in Figure~\ref{fig:introduction}, reference answers and model outputs often express the same numerical value using different formulas.
To enable robust verification, a reliable verifier must be capable of performing arithmetic transformations before making a judgment.
In this paper, we propose to equip the verifier with external tools such as a Python interpreter.
Therefore, our proposed \model can perform precise computations on both formulas to derive aligned numerical results, thereby ensuring accurate verification.

\begin{figure}[t]
        \centering
	\includegraphics[width=0.6\linewidth]{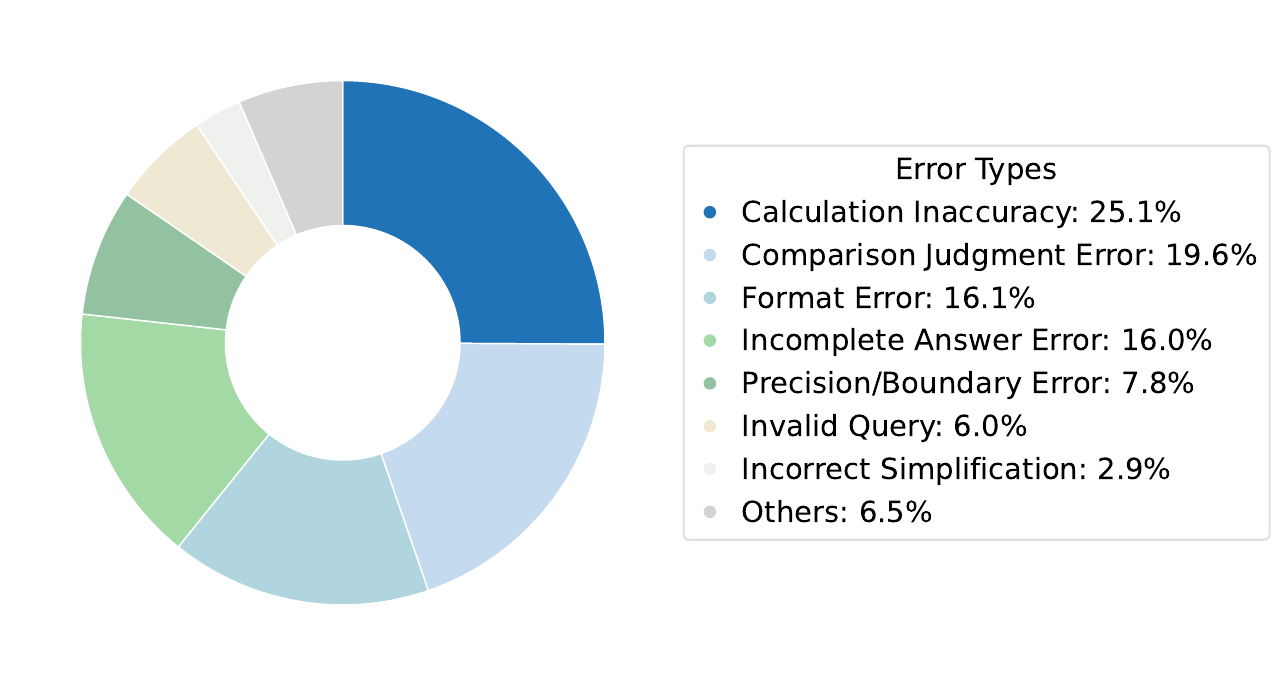}
        \vspace{-10pt} 
        \caption{Error distribution on existing verification methods. More Details can be found in the Appendix~\ref{appendix:error_analysis}.}
 \label{fig:error_types}
 \vspace{-10pt}
\end{figure}

However, foundation models are not well-tuned on precise tool-augmented verification, which may issue inaccurate and redundant tool calls during the process.
To address this challenge, we propose a novel verifier training methodology. We begin by collecting a diverse set of questions spanning multiple domains, including mathematics, natural sciences, and logical reasoning.
We first conduct a comprehensive analysis of how common verification methods fail on these questions, identifying typical failure modes in such settings.
As illustrated in Figure~\ref{fig:error_types}, we identify two major challenges:
(i) Calculation inaccuracy represents the largest proportion of verification failures, highlighting the limitations of existing verifiers in computation-intensive contexts.
(ii) Error types are diverse and follow a long-tail distribution, making it difficult for previous data augmentation approaches to achieve comprehensive coverage.
Based on these insights, we design an automatic method for constructing verifier training data that synthesizes challenging computation-heavy instances. 
This approach increases both the density of queries prone to verification errors and the proportion of cases that require computational tool-augmented reasoning.

After constructing this diverse and challenging training dataset, we propose a two-stage training strategy to equip the verifier with effective tool-use capabilities in computation-heavy scenarios: (i) cold-start supervised fine-tuning, followed by (ii) reinforcement learning with tool-call encouragement.
The rationale for this approach is that while the cold-start phase enables the model to invoke tools, such invocations remain unreliable—only 76\% of tool-invoked verifications are initially correct.
To address this issue, the second stage employs agentic reinforcement learning with rewards designed to explicitly encourage precise and effective tool usage, which boosts tool-invoked verification accuracy from 76\% to 90\%.

To summarize, our contributions are threefold: 
\begin{itemize}
  \item We propose \textbf{\model}, a tool-augmented answer verification model. By invoking external tools (e.g., Python interpreter), it performs precise computation and algebraic simplification to assess numeric equivalence between scientific formulas.  

  \item We develop a two-stage training framework consisting of cold-start fine-tuning followed by a reinforcement learning stage that encourages effective tool use, substantially increasing the correctness of tool-invoked verifications on computation-heavy questions.

  \item Experimental results demonstrate the effectiveness of \model in terms of both verification performance and downstream RLVR applications. \model achieves state-of-the-art results on VerifyBench and Sci-Bench. As a fast and accurate reward model in RLVR, it also consistently outperforms rubric-based and CoT-only verifiers on both AIME'24 and AIME'25.
  
\end{itemize}

\section{Related Works}
\label{sec:related_work}


\subsection{Answer Verification}
Answer verification is the automated process of deciding whether a candidate response is factually correct to the provided reference answer. Significant progress has been made in the development of LLM‑output verification\cite{liu2025inference, xia2025agentrm}, which can broadly be divided into two main classes: rule-based verification and  model-based verification.Early rule‑based methods like Math-Verify\cite{Kydlicek_Math-Verify_Math_Verification} extract the final answer via heuristics and compare it against the ground truth using predefined rules. However, these approaches struggle with unstructured model outputs and tend to incur a high false‑negative rate\cite{huang2025pitfalls}. To address these shortcomings, model‑based verifiers exploit the flexibility of LLMs to interpret and evaluate unpredictable output. Xverify\cite{chen2025xverify} and CompasVerifier\cite{liu2025compassverifier} are designed to provide efficient and robust answer verification by carefully constructing training examples. While effective, they are constrained by the verifier’s inherent reasoning and calculation competency—particularly in computation‑intensive math and unit‑sensitive STEM problems. 

In this paper, we concentrate on the model-verifiaction reward setting , notably in calculate-intensive and science domains where existing verifiers usually fail. To overcome these limitations, we propose CosineVerifier, which endows the model with external tool capabilities (e.g., a Python interpreter and a unit‑conversion toolkit) to reliably validate diverse LLM outputs while delivering efficient, robust and trustworthy verdicts. 

\subsection{Tool Integrated Reasoning}
A growing line of work investigates improving LLM reasoning by augmenting models with external tools, yielding consistent gains in accuracy, calibration, and sample efficiency across math, program, and multi-turn QA tasks\cite{mai2025agentrl, chen2025r1ci, dong2025toolstar, zhang2025nemotronn1, ma2025tooltrain, lin2025understandingtir}. ReTool\cite{feng2025retool}, ToRL\cite{li2025torl} and SimpleTIR\cite{xue2025simpletir} train models to interleave real-time code execution with language reasoning and to learn when/how to call tools via reward signals, yielding strong gains on math-reasoning benchmarks. VTool-R1\cite{wu2025vtoolr1} integrates Python-based visual editing tools (box, mask, highlight) into the RFT loop so the model learns to interleave intermediate visual steps with textual chain-of-thought. Extending beyond code, Search-R1\cite{jin2025searchr1} casts web search as an action space and trains query generation/revision via outcome rewards for multi-turn retrieval-augmented reasoning. Taken together, these approaches recast reasoning over a unified tool-action space, spanning code, vision, and search—where outcome-level rewards teach models when to plan, execute, and reflect, yielding stronger generalization and reliability beyond base models.

To the best of our knowledge, CosineVerifier is the first outcome answer verifier that augments LLM-based judgments with executable tools, enabling complex unit-consistent calculations, tolerance-aware numeric checks and robust string format equivalence for general verification process.

\section{Preliminaries}
\label{sec:preliminaries}
\paragraph{Answer verification} We consider the task of outcome reward verification as follows: given a set of questions $Q=\{q_1, q_2,\dots, q_n\}$, their targeted model-generated answers $A = \{a_1, a_2,\dots,a_n\}$ and their corresponding reference answers $A_{ref} = \{a_{ref}^{1}, a_{ref}^{1}, \dots, a_{ref}^{n}\}$, the goal is to evaluate the correctness of the model-generated answers $A$, which can be formulated as:
\begin{equation}
    R = V_{\theta}(Q, A, A_{ref})
\end{equation}
where $R=\{r_1,r_2,\dots,r_n\} $ with $r\in\{0,1\}$ is the set of binary verification results and $V_{\theta}$ is the verification model parameterized by $\theta$.

\section{Methodology: \model}
\label{sec:method}

\begin{figure}[H]
        \centering
	\includegraphics[width=1.0\linewidth]{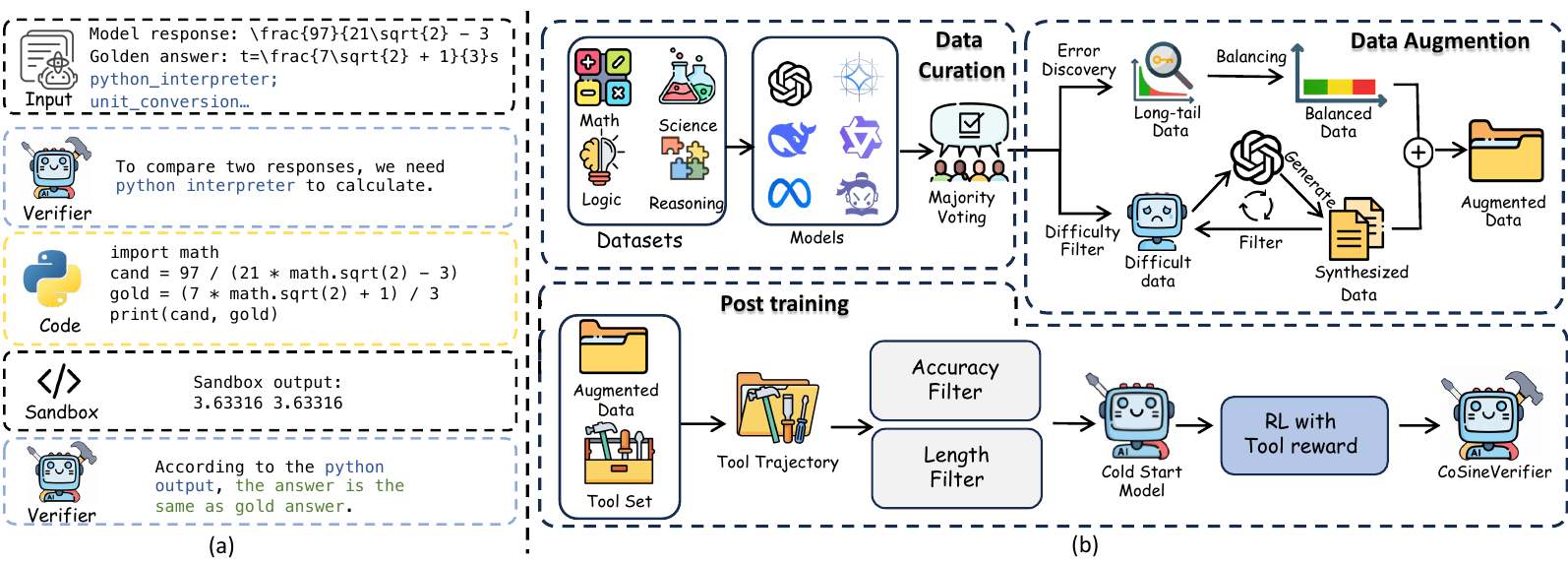}
        \caption{(a) Workflow of tool-augmented verification. (b) The overall training framework of \model.}
 \label{fig:main}
 \vspace{-10pt}
\end{figure}

In this section, we outline the workflow and training procedure of \model, which is endowed with agentic tool-use capabilities for accurate outcome verification. 
Figure~\ref{fig:main} shows the overall framework of \model.
Section~\ref{sec:tool-augmented verification} describes how our \model interleaves reasoning traces with dynamic tool execution to reach a final verdict. Section~\ref{sec:data collection} details our large-scale, multi-domain data collection and annotation pipeline, Section~\ref{sec:data augmentation} then describes our data augmentation strategy on long-tail and difficult data. 
Finally, Section~\ref{sec:post-training} presents the training recipe of \model, including cold-start supervised fine-tuning and reinforcement learning.

\subsection{Tool-augmented Verification}
\label{sec:tool-augmented verification}

To facilitate the reliable outcome verification, we equip our verification model with external tools that can dynamically execute computations and norliaze units, enabling precise judgments on calculation-intensive and science tasks. Specifically, as shown in Figure \ref{fig:main} (a), when facing a verification problem, our \model-4B-Tool first decides whether to use an external tool. If external tool call is needed, we parse the tool arguments(e.g. python code) and send them to sandbox environments. After the sandbox environment completes the execution and returns the output, \model-4B-Tool gives its final judgement based on the sandbox output. Full case study can be found in Appendix \ref{appendix:case_study}.

\subsection{Data Curation} \label{sec:data collection}
\label{sec:qa_acquisition}
\paragraph{Data Collection} Answer verification in real-world applications is challenging for two reasons: (i) the  diversity of questions encountered in practice, and (ii) the variability of model outputs (format, length, and reasoning style). To address these challenges, we construct a large-scale candidate training data, comprising 1.14 million samples. These samples were collected from 15 datasets using 14 large language models (LLMs), spanning mathematics, science, logical reasoning, and commonsense knowledge domains. This provides broad coverage for training robust verification models. In addition, to strengthen CosineVerifier on practical verification tasks, we deliberately go beyond static benchmarks (in contrast to prior collections such as CompassVerifier\cite{liu2025compassverifier} and Xverify\cite{chen2025xverify}). A substantial portion of our data is sampled from competition-level math and science questions, where verification often fails in practice. During the construction of of our data, we also ensure cross-domain coverage, and to further increase robustness, we also prepend different prompt prefixes(e.g. Let's think step by step) on a subset of training data, promoting generalization to diverse prompting styles.

\paragraph{Data Annotation} After constructing a large and diverse training data, we introduce an iterative annotation pipeline that produces reliable verification labels. We first employ a set of models (Compass-Verifier-32B, qwen3-4b-2507, and Qwen3-32B), with each model providing three independent annotations for a total of nine per sample. We then retain 63,714 samples that show any disagreement among these nine annotations. For samples that showcase high disagreement rate(defined as more than three out of nine annotations differing), we further send them to a more powerful model(GPT-o3) with majority voting, ensuring high label fidelity on harder questions.

\subsection{Data augmentation} 

\label{sec:data augmentation}
\paragraph{Long-tail data augmentation}
To understand when the verifier performs poorly, we conducted a discrepancy analysis over 63{,}374 disagreement cases. We first prompted Qwen3-32B to produce an error category for each sample and then leverage GPT-o3 to summarize these categories into 15 error types. As shown in Figure \ref{fig:error_types}, all error types reveal a pronounced skew: the top five categories—dominated by Calculation Inaccuracy and Exact Match Failure, account for over 85\% of errors, while the remaining ten categories are sparsely represented. This distribution makes low-frequency errors harder to learn. To counter this imbalance, we use gpt-o3 to synthesize targeted long-tail verification examples under constrained generation settings, yielding 10k long-tail synthesize instances. We 
further incorporate these samples into training for both the labeling verifier and the tool-augmented verifier, improving coverage of rare error modes. All training data statistic can be found in Appendix \ref{appendix:training details}.

\paragraph{Difficulty-oriented data augmentation}After cold-start, our model acquired foundational tool-use and verification skills. To further amplify its performance, we focused on curating a high-difficulty post-training dataset using an iterative self-instruct framework. In each round, GPT-o3 synthesized new questions conditioned on the current hard cases. Our cold-start model then attempted these questions. We only retrained incorrect instances, and merged these instances with the reference set for the next iteration. After four such iterations, the procedure produced a reinforcement-learning corpus of 9,456 examples. 
A detailed algorithm is depicted in \ref{alg:qs}. 

\begin{algorithm}[t]
\DontPrintSemicolon
\SetAlgoLined
\caption{Iterative Difficult Data Synthesis}
\label{alg:qs}
\KwIn{Initial hard set $\mathcal{D}_0$; questioner $Q$; solver $S$; rounds $T$.}
\KwOut{Reinforcement-learning corpus $\mathcal{R}$.}

$\mathcal{R} \gets \{\}$\;
$\mathcal{D} \gets \mathcal{D}_0$

\For{$t \gets 1$ \KwTo $T$}{
  $\mathcal{Q}_t \gets Q(\text{conditioned on }\mathcal{D})$\;
  \tcp{Conditioned generation by Questioner}
  $\mathcal{F}_t \gets \{\,x \in \mathcal{Q}_t \mid S(x)\ \text{is incorrect}\,\}$\;
  \tcp{Accumulate failures as next-round context}
  $\mathcal{R} \gets \mathcal{R} \cup \mathcal{F}_t$\;
  $\mathcal{D} \gets \mathcal{D} \cup \mathcal{F}_t$\;
}
\Return $\mathcal{R}$\;
\end{algorithm}

\subsection{Post Training} \label{sec:post-training}
\subsubsection{Cold-Start}
To ensure verification accuracy and robustness, we equip CosineVerifier with two executable tools: a Python interpreter and a unit-conversion utility. During the preliminary experiment of \model, we find that directly apply reinforcement learning on base model yields performance degradation, due to the base model's limited tool-calling capabilities. We therefore implement a cold-start stage to equip the model with foundational tool-use proficiency. Specifically, we first collect real-world tool-use traces from our training data and retain only those cases that (i) produce a correct final answer and (ii) contain explicit tool-call trajectories. To boost the efficiency of \model , we discard model internal \texttt{<think>}…\texttt{</think>} content, and further filter out responses longer than 200 tokens to keep the cold-start corpus compact and efficient. This procedure yields 29{,}339 tool-usage examples used to bootstrap training.

During training \model-4B-Tool, we mask the loss on tool-execution outputs (e.g., interpreter traces and unit-conversion results), which prevents the model from merely imitating tool dumps and improves optimization stability. We also train \model-Label series models, which are train on the full 63,374 disagreement data together with 10k model-generated long-tail instances, and providing verdict as binary classification with targets \textit{Correct} and \textit{Incorrect}.

\subsubsection{Reinforcement Learning}
\paragraph{Overall framework} For \model, we also undergone reinforcement learning to incentivize more reasoning ability. Specifically, we adopt the DAPO~\cite{yu2025dapo} algorithm, which is an improved variant of GRPO\cite{shao2024deepseekmath}. Given a prompt–answer pair $(q,a)$, we draw $G$ rollouts $\{s_i\}_{i=1}^{G}$ from the behavior policy $\pi_{\theta_{\text{old}}}$. With access to the grounded answer $a$, each rollout receives a scalar reward $R_i = R(s_i,a)$ and we then optimize the current policy $\pi_{\theta}$ using the clipped policy-gradient objective below:
\begin{equation}
\begin{aligned}
\mathcal{J}_{\text{DAPO}}(\theta) =\quad& \mathbb{E}_{(q,a)\sim \mathcal{D}, \{o_i\}_{i=1}^G\sim \pi_{\theta_\text{old}}(\cdot\mid q)}\\&
\Bigg[\frac{1}{\sum_{i=1}^{G}|o_i|}\sum_{i=1}^{G}\sum_{t=1}^{|o_i|} 
\min \Big( r_{i,t}(\theta) \hat{A}_{i,t},  
\ \text{clip} \Big( r_{i,t}(\theta), 1 - {\varepsilon_{\text{low}}}, 1 + {\varepsilon_{\text{high}}} \Big) \hat{A}_{i,t} \Big) \Bigg]
\\
\text{s.t.}\quad& 0< \Big|\{o_i\mid\texttt{is\_equivalent}(a,o_i)\}\Big|< G,
\label{eq:dapoloss}
\end{aligned}
\end{equation}
where
\begin{equation}
    r_{i,t}(\theta)=\frac{\pi_{\theta}(o_{i,t} \mid q, o_{i,<t})}{\pi_{\theta_{\text{old}}}(o_{i,t} \mid q,o_{i,<t})},\quad\hat{A}_{i,t} = \frac{R_i - \text{mean}(\{R_i\}_{i=1}^G)}{\text{std}(\{R_i\}_{i=1}^G)}.
\label{eq:advantage_calculation}
\end{equation}

\paragraph{Reward design} We decompose the \model's outcome-based reward into two terms. The first is an answer-correctness component:
\[
R_{\text{ans}}=\mathbb{I}\{\,\hat{y}=y\,\},
\]
which assigns 1 when the model’s prediction $\hat{y}$ matches the reference $y$, and 0 otherwise. 

To promote tool use during training, we introduce a tool-use encourage reward. 
Let $u$ denote the number of tool invocations in the trajectory. We award a bonus when the model is correct and uses an external tool, and impose a penalty when the model is incorrect without relying on function calling:
\[
R_{t}=
\begin{cases}
0.5, & \text{if }\mathbb{I}\{\,\hat{y}=y\,\}=1 \text{ and } u>0,\\
-0.5, & \text{if }\mathbb{I}\{\,\hat{y}=y\,\}=0 \text{ and } u=0,\\
0, & \text{otherwise}.
\end{cases}
\]
The final reward is the sum of the correctness and tool-encourage terms:
\[
R = R_{\text{ans}} + R_{t}.
\]
This shaping explicitly encourages effective tool use trajectories that leverage tools to reach correct answers and discourages failures that eschew tools.

\section{Experiments}
\label{sec:experiment}
\begin{table}[t]
\centering
\centering
\setlength{\tabcolsep}{6pt}
\resizebox{\linewidth}{!}{%
\begin{tabular}{l|c|c|c|c}
\toprule
\textbf{Model} & \textbf{VerifyBench} & \textbf{VerifyBench(Hard)} & \textbf{SCI-Bench} & \textbf{Avg. Tokens} \\
\midrule
\rowcolor{gray!10}
\multicolumn{5}{c}{\textbf{CoT Verifier}}\\
\midrule
\multicolumn{5}{l}{\small\textbf{Closed-source Models}\rule[-0.5ex]{0pt}{1.6ex}}\\
o3\cite{openai2025o3}                     & \underline{96.1} & \underline{88.7} & \underline{87.5} & 206.7 \\
GPT-4o\cite{openai2024gpt4o}                  & 96.0 & 84.6 & 86.0 & 192.4 \\
Gemini2.5-Flash\cite{comanici2025gemini}         & 96.0 & 86.0 & 85.9 & 193.0 \\
\midrule
\multicolumn{5}{l}{\small\textbf{Open-source Models}\rule[-0.5ex]{0pt}{1.6ex}}\\

GPT-oss-20B\cite{agarwal2025gpt}             & 92.2 & 84.7 & 85.0 & 221.0 \\
LLaMA3.3-70B-Instruct\cite{dubey2024llama}   & 94.8 & 77.2 & 84.8 & 347.3 \\
Qwen3-4B\cite{yang2025qwen3}                & 92.6 & 80.3 & 82.0 & 1156.7 \\
Qwen3-8B\cite{yang2025qwen3}                & 93.7 & 83.6 & 83.9 & 926.6 \\
Qwen3-32B\cite{yang2025qwen3}               & 94.7 & 85.2 & 83.5 & 798.8 \\
Qwen3-4B-Instruct-2507\cite{yang2025qwen3}  & 94.7 & 84.1 & 82.4 & 869.7 \\
Qwen3-235B-A22B-2507\cite{yang2025qwen3}    & 94.4 & 87.7 & 82.6 & 1885.3 \\
CompassVerifier-7B (CoT)\cite{liu2025compassverifier} & 93.5 & 82.6 & 84.2 & 234.7 \\
CompassVerifier-32B (CoT)\cite{liu2025compassverifier} & 95.9 & 86.5 & 85.5 & 213.0 \\
\rowcolor{cyan!7}
CoSineVerifier-Tool-4B & \textbf{96.6} & \textbf{91.9} & \textbf{89.7} & \textbf{95.3} \\
\midrule
\rowcolor{gray!10}
\multicolumn{5}{c}{\textbf{Labeling Verifier}}\\
\midrule
XVerify-8B-I\cite{chen2025xverify}            & 92.5 & 83.3 & 78.1 & 1.0 \\
CompassVerifier-7B\cite{liu2025compassverifier}      & 93.5 & 85.2 & 85.7 & 1.0 \\
CompassVerifier-32B\cite{liu2025compassverifier}     & \textbf{96.3} & \underline{88.9} & 85.3 & 1.0 \\
\rowcolor{cyan!7}
CoSineVerifier-4B       & \underline{95.7} & 85.4 & \underline{85.9} & 1.0 \\
\rowcolor{cyan!7}
\textbf{CoSineVerifier-32B}  & \underline{95.7} & \textbf{90.0} & \textbf{86.4} & \textbf{1.0} \\
\bottomrule
\end{tabular}%
}
\par\rule{0pt}{10pt} 
\caption{Main results on three verify benchmarks. We separate CoT verifier and Labeling verifier for fair comparison, where CoT verifier output a reasoning chain before achieving final answer while labeling verifier directly gives it's verdict. \textbf{Bold} highlights the overall best accuracy, while \underline{underline} highlights the second best accuracy. We report mean@3 accuracy and average token used per question.}
\label{tab:main-verifier-results}
\vspace{-15pt}
\end{table}

\subsection{Experiment Setups} \label{subsec: exp setup}
\paragraph{Datasets and Baselines} We evaluate the \model family on three benchmarks: VerifyBench, VerifyBench-Hard\cite{yan2025verifybench}, and Sci-Bench~\cite{zheng2025sci}, which are challenging verification benchmarks spanning math, science, commonsense knowledge and general reasoning domains with 2000, 1000 and 2500 samples.  Our baselines mainly include two kinds of models: labeling verifier and COT verifiers, with difference on whether contain reasoning chains for verification and we report accuracy as mean@3 and efficiency as average output tokens per verdict on these benchmarks.

\paragraph{Implementation details} 
Our CoSinVerifier-4B-Tool, CosineVerifier-4B and CoSinVerifier-32B are trained from Qwen3-4B-Instruct-2507, Qwen3-4B, and Qwen3-32B\cite{yang2025qwen3}. We use OpenRLHF\cite{hu2024openrlhf} for supervised fine-tuning and verl\cite{sheng2025hybridflow} framework for reinforcement learning. All models are trained for one epoch on 32xA800 80G GPUS during supervised finetuning with 1e-5 learning rate, and 60 step for reinforcement learning with 1e-6 learning rate. All training details are listed in Appendix \ref{appendix:training details}

\subsection{Main Results}
\paragraph{\model series achieve strong verification accuracy compared to baselines} As shown in Table \ref{tab:main-verifier-results}, our \model series outperforms all baselines across three verify benchmarks. Specifically, for CoT Verifier, our \model-Tool-4B surpasses baselines by 0.5\%-14.7\% with noticeably less tokens, especially on harder benchmarks like VerifyBench(hard) and Sci-Bench, where 
complex calculation and long-context string comparing is needed to verify the LLM-generated answer. In addition, for labeing verifier, \model-4B and \model-32B  also surpass other labeling verifiers by 
0.7\%-1.1\%, demonstrating stronger verification skills under single-token scenarios.

\begin{figure}[t]
    \centering
    \hspace{-3cm}
    \begin{subfigure}[b]{0.55\linewidth} 
        \centering
        \includegraphics[height=5.5cm, keepaspectratio]{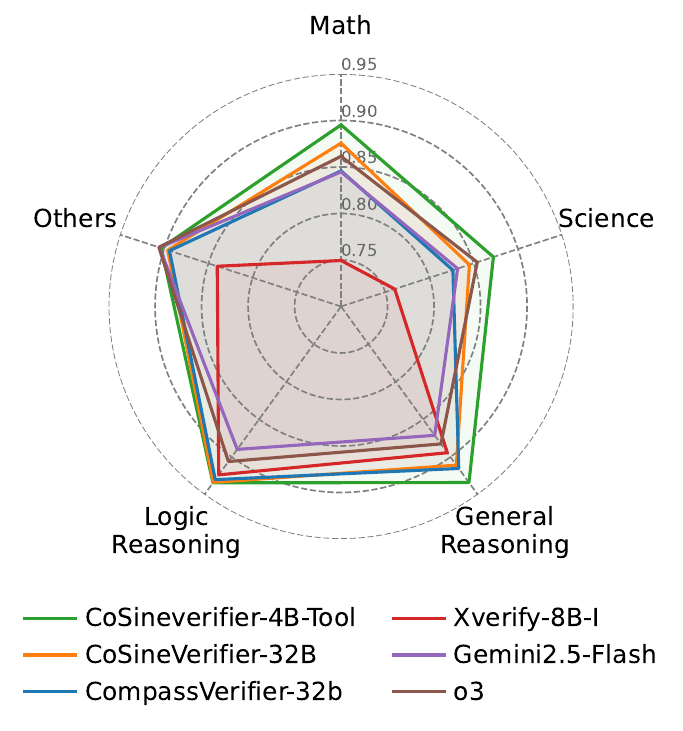}
        \caption{}
        \label{fig:radar_analysis}
    \end{subfigure}
    \hspace{-0cm}
    \begin{subfigure}[b]{0.4\linewidth} 
        \centering
        \includegraphics[height=4.8cm, keepaspectratio]{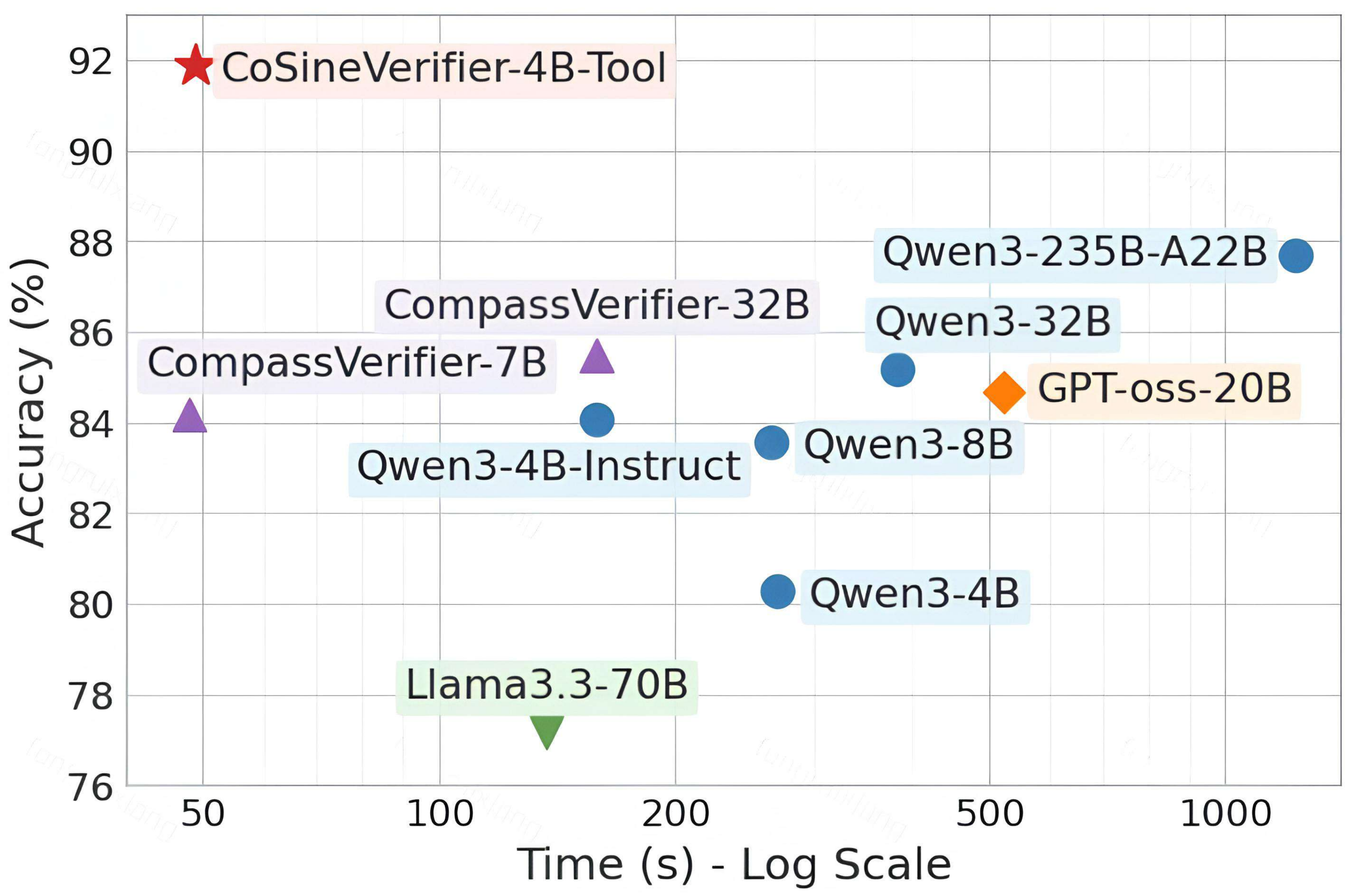}
        \captionsetup{margin={2.0cm, 0pt}} 
        \caption{}
        \label{fig:time_analysis}
    \end{subfigure}
    
    \caption{Performance Analysis. (a) Analysis on accuracy across different domains. (b) Inference efficiency analysis of \model-4B-Tool.}
    \label{fig:analysis_comparison}
    \vspace{-8pt}
\end{figure}

\paragraph{Small models with external tools excel large models} Figure \ref{fig:analysis_comparison} demonstrates the performance comparison of our \model against various model across different domains. Among three benchmarks, even close-source SOTA models like o3 still struggle with reliable verification on compute-intensive and scientific reasoning tasks. In contrast, both \model-4B-Tool and \model-32B demonstrate exceptional proficiency in the Math and Science domains while achieve competitive or even superior performance on the remaining domains. Despite being much more smaller on parameter size,  the superior accuracy exhibited by \model-4B-Tool validates the efficacy of integrating external calculation tools to enhance the model's verification capabilities. These results demonstrate the feasibility of tool-augmented, small-scale models to exceed the performance of large SOTA models in complex verification tasks.
\paragraph{\model-4B-Tool shows superior efficiency while maintaining high accuracy} We further evaluate the efficiency of \model-4B-Tool against several baselines in Figure \ref{fig:analysis_comparison}. To ensure a fair comparison, all models were benchmarked on 4xNVIDIA A800 GPUs. We report the mean inference latency computed over 1000 randomly sampled prompts. While \model-4B-Tool achieves high accuracy, incorporating external tools does not incur significant inference latency. We attribute this efficiency to two primary factors: (1) our efficiecy-driven cold-start data construction methodology, which explicitly optimizes for short and precise tool-calling trajectories, resulting in 2.4x-19x times less token usage compared to other models. (2) the computational efficiency afforded by the model's compact 4B parameter count, which minimizes the overhead of the backbone model during inference.



\subsection{Ablation Study}
\begin{figure}[t]
        \centering
	\includegraphics[width=0.7\linewidth]{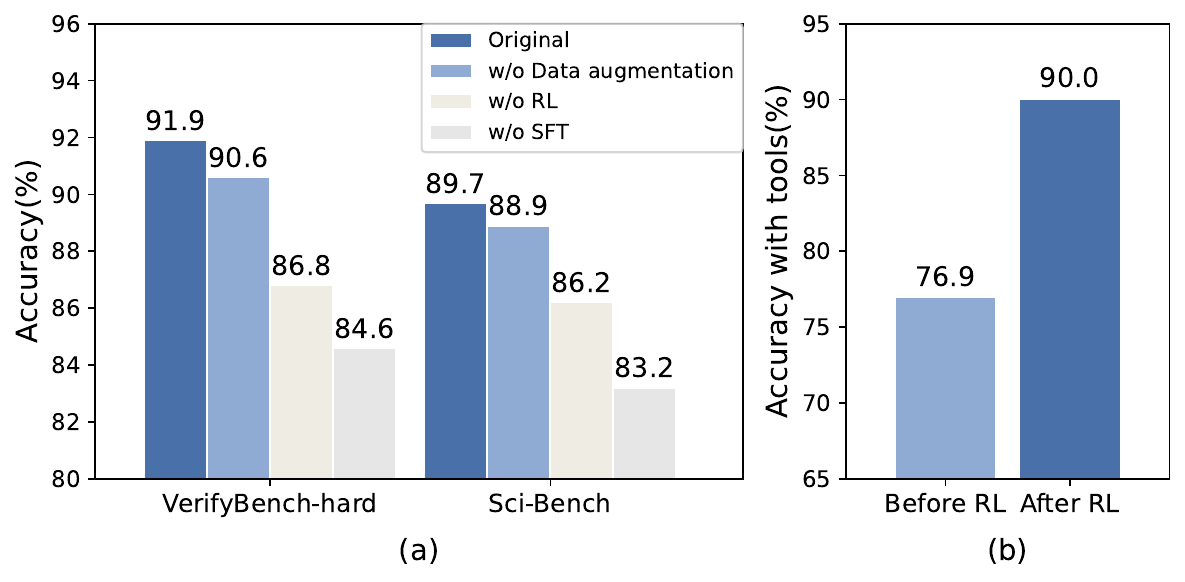}
        \caption{(a) Ablation study of \model-4B-Tool. (b) Average verification accuracy with tools before RL and after RL}
 \label{fig:ablation}
 \vspace{-15pt}
\end{figure}

\paragraph{Two stage training is crucial for \model-4B-Tool} As shown in Figure \ref{fig:ablation}(a), we conduct ablation studies on each module's contribution to our two-stage training framework. We first observe that after removing cold start stage for \model-4B-Tool, the model's performance drop sharply. We then discover this is because model without cold-start fails to conduct accurate function calling and often becomes trapped in tool-use loops, harming performance while significantly increase inference latency. In addition, model trained without reinforcement learning(RL) also demonstrate weaker verification performance. To investigate this, we also analyze the accuracy on samples that have tool trajectories before and after RL stage.  As shown in Figure \ref{fig:ablation}(b), the accuracy on samples with tools improved markedly from 76.9\% to 90.0\%. This result indicates that RL teaches the model how to leverage tools for accurate verification, rather than merely to initiate a tool call. In conclusion, each stage of our two-stage framework is crucial for robust verifier training. Together, they equip the model to determine precisely when and how to use tool calls for verification.

\paragraph{Data augmentation can further boost performance} As shown in Figure\ref{fig:ablation}(a), removing data augmentation also yields inferior performance. We attribute this to the fact that verification tasks often contain rare but difficult samples, such as truncated responses, multiple self-reflections, and meaningless repetitions. These samples are relatively more difficult to verify and are underrepresented in ablation training data, preventing the model from learning to handle them as effectively as common verification samples. Incorporating our data augmentation methods can exposing the model to a more balanced and challenging training distribution, which ensure the our model have high verification accuracy in rare and difficult circumstances.

\section{Application in RLVR}
The primary objective of our \model series is to establish a reliable and efficient outcome reward verifier tailored for real-world data processing and reinforcement learning. To avoid potential biases inherent in existing benchmarks and validate the efficacy of \model as a reward model with RL training, we further evaluate answer-verification methods in RLVR settings. Specifically, utilizing an on-policy GRPO algorithm, we train Qwen3-4B-Instruct-2507 on competition-math problems with 42K training data drawn from DAPO-Math-17k\cite{yu2025dapo}, OpenR1-Math-220k\cite{openr1}, and DeepScaleR-Preview\cite{deepscaler2025}. We compare our approach against various verifiers, including both rule-based and model-based methods, under identical training configurations. We report mean@32 accuracy on challenging math benchmarks AIME 2024 and AIME 2025 and comprehensive training details are provided in Appendix \ref{appendix:training details}.
\vspace{-15pt}
 \begin{figure}[H]
        \centering
	\includegraphics[width=1\linewidth]{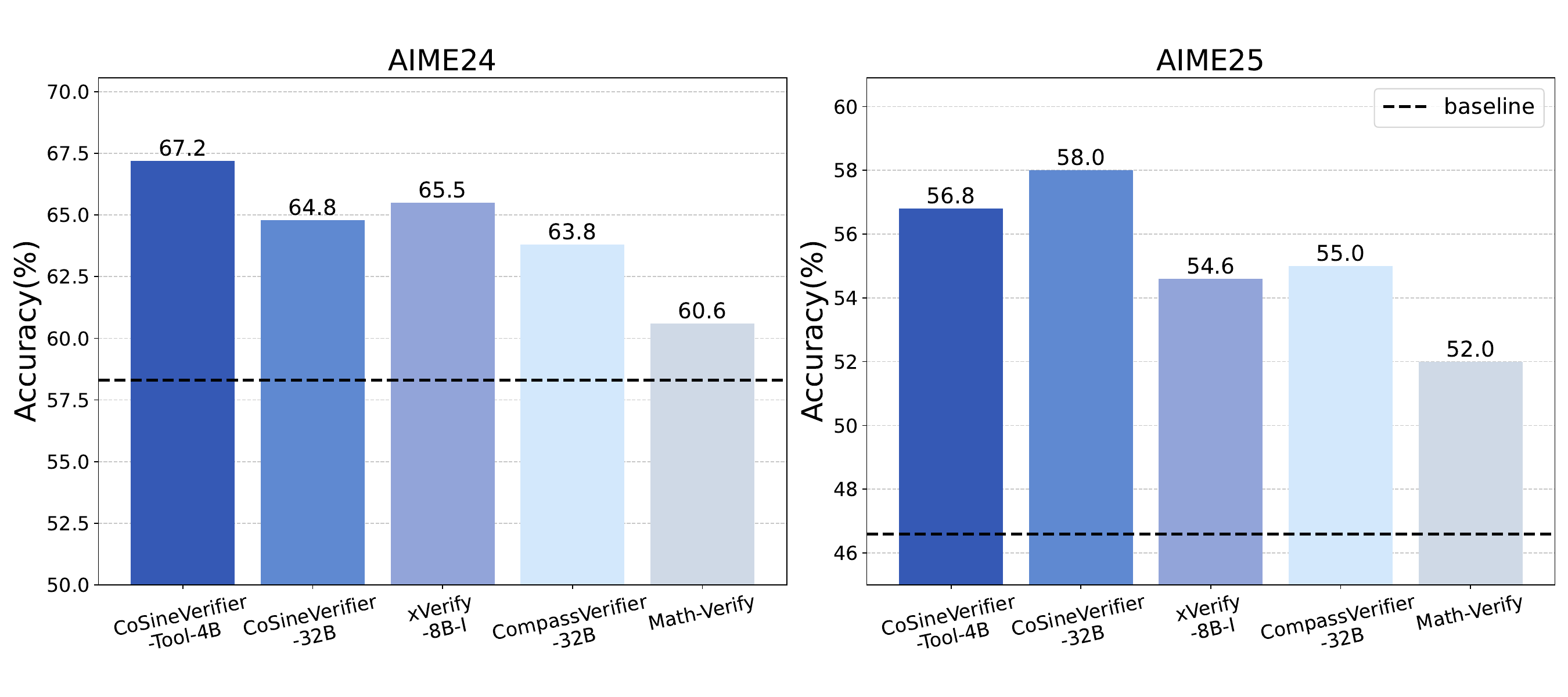}
        \caption{Results of using \model as a reward model on AIME24 and AIME25 benchmark}
 \label{fig:reward model}
 \vspace{-8pt}
\end{figure}

\paragraph{RL training result} As illustrated in Figure \ref{fig:reward model}, training with our \model enables the base model to achieve stronger reasoning capabilities, yielding improvements of 1.7\%--6.6\% on AIME 2024 and 1.8\%--6.0\% on AIME 2025. Moreover, a clear gap exists between rule-based and model-based verifiers. This highlights that static methods alone are inadequate for mapping diverse LLM outputs to reference answers. We believe a precise and reliable reward signal is beneficial for stable RL training, especially when the model is already well-tuned and requires more challenging RL training data to surpass its reasoning ceiling.


\begin{figure}[t]
        \centering
	\includegraphics[width=0.5\linewidth]{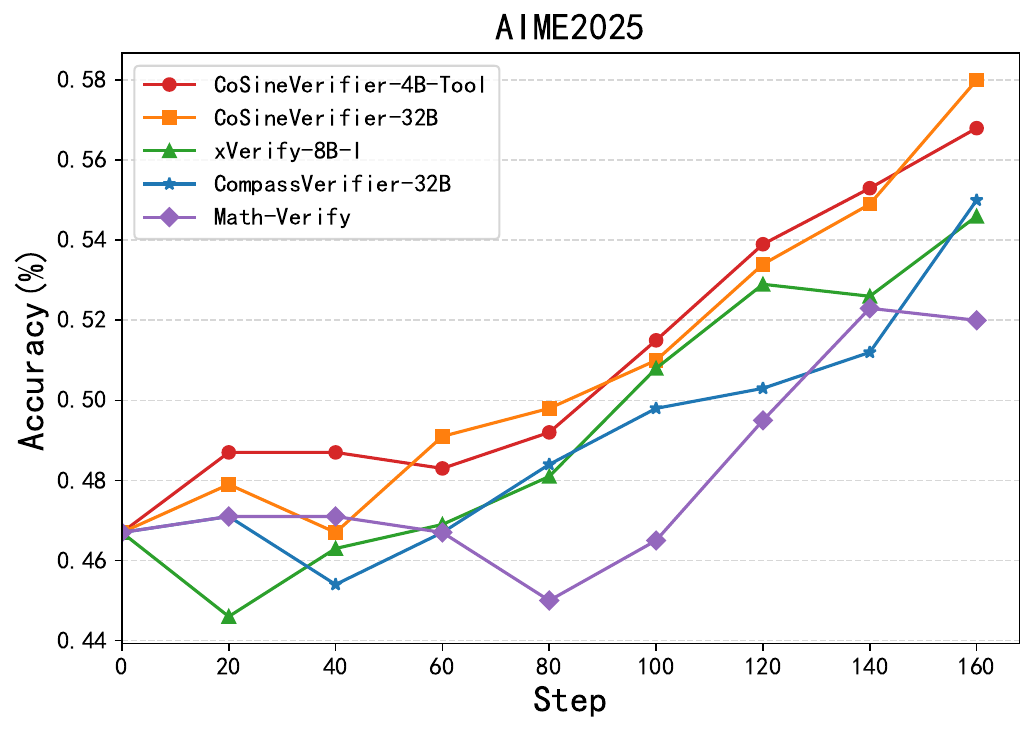}
        \caption{Accuracy of AIME 2025 during training process}
 \label{fig:aime25_step}
 \vspace{-10pt}
\end{figure}

\paragraph{From the perspective of training process} We also dive into the accuracy change trend on AIME 2025 during RL training process. We aim to answer the following question: what is the impact of using different verifiers on downstream task performance under identical RL training settings? As shown in Figure \ref{fig:aime25_step}, our \model consistently yields a clear lead over all baselines, demonstrating that base model can benefit significantly from optimizing verifiers during RL training. In contrast, weaker verifier like Math-Verify perform worse than \model series, demonstrating a direct correlation between a verifier's capability and base model's reasoning performance during reinforcement learning. As recent research increasingly utilizes challenging datasets that elicit highly complex model responses and reference answers\cite{phan2025humanity,guan2025rstar}, this trend naturally motivates us to develop a more reliable verification system that ensures robust and accurate reward signaling to boost the performance of base models.

\section{Conclusion}
In this work, we introduce the \model series, a family of answer verifiers designed for computation-oriented scientific questions. Our core contribution, CoSineVerifier-Tool-4B, demonstrates that a compact 4B model, when augmented with external tools like a Python interpreter, can significantly outperform much larger models on challenging verification benchmarks. This tool-integrated approach enables the precise evaluation of algebraic equivalences and physical constant alignments, addressing a key weakness of existing LLM-based verifiers. Furthermore, when applied as a reward model in RLVR tasks, \model series yields clear performance improvements on AIME24 and AIME25, highlighting the importance of accurate, tool-enhanced reward signals for training more capable models.

\bibliographystyle{unsrt}  
\bibliography{references}

\newpage
\appendix
\section{Error Analysis}
\label{appendix:error_analysis}
\begin{table}[h]
\centering
\begin{tabular}{@{}l p{8.5cm} r@{}}
\toprule
\textbf{Error Name} & \textbf{Description} & \textbf{Percentage} \\ 
\midrule
Calculation Inaccuracy & The final numerical value is mathematically incorrect. & 25.089\% \\
Comparison Judgment Error & The response (e.g., a sequence) does not perfectly match the reference. & 19.618\% \\
Format Error & The answer's structure violates the required output format. & 16.085\% \\
Incomplete Answer Error & Fails to provide all required answers, e.g., missing candidates or sub-parts. & 15.957\% \\
Precision/Boundary Error & Incorrect rounding or definitions of interval endpoints, e.g., open/closed. & 7.843\% \\
Invalid Query & The error originates from a defective question or reference answer. & 6.007\% \\
Incorrect Simplification & Fails to simplify to minimal form or simplifies erroneously. & 2.938\% \\
Constraint Violation & The answer ignores a specific rule, e.g., ``use each number once''. & 2.166\% \\
Missing Final Result & Provides only reasoning or code without the conclusive answer. & 1.836\% \\
Refusal or Inconclusive Response & Fails to provide a clear answer or states it is unanswerable. & 1.355\% \\
Extraneous Content Error & The response includes correct data mixed with unrelated, incorrect information. & 0.677\% \\
Self-Correction Failure & The model initially answers correctly but changes to an incorrect answer. & 0.285\% \\
Truncated Response & The response is clearly cut off and incomplete. & 0.085\% \\
Garbled or Corrupted Output & The response contains unreadable characters or significant noise. & 0.046\% \\
Meaningless Repetition Error & The response repeats text extensively, hiding any valid answer. & 0.013\% \\ 
\bottomrule
\end{tabular}
\vspace{0.2cm}
\caption{Error Types and Distributions}
\end{table}

\section{Training Data Statistics}
In this section, we provide the relevant statistics of our \model training data.

\begin{table}[ht]
\centering
\begin{tabular}{lrr}
\toprule
\textbf{Source} & \textbf{Count} & \textbf{Rate} \\
\midrule
Science & 452661 & 39.70\% \\
Math & 271415 & 23.80\% \\
Drop & 103070 & 9.04\% \\
SuperGPQA & 60356 & 5.29\% \\
Ceval & 59501 & 5.21\% \\
Cmmlu & 52222 & 4.58\% \\
MMLU\_pro & 50301 & 4.41\% \\
ARC & 33148 & 2.90\% \\
SimpleQA & 18103 & 1.58\% \\
ChineseSimpleQA & 12720 & 1.11\% \\
BBH & 11794 & 1.03\% \\
Korbench & 9561 & 0.83\% \\
GAOKAOBench & 3281 & 0.28\% \\
GPQA & 1913 & 0.16\% \\
\midrule %
\textbf{Total} & \textbf{1,140,046} & \textbf{100.00\%} \\
\bottomrule
\end{tabular}
\vspace{0.2cm}
\caption{Candidate training data distribution}
\label{tab:benchmark_allocation}
\end{table}

\begin{table}[ht]
\centering
\begin{tabular}{lrr}
\toprule
\textbf{Source} & \textbf{Count} & \textbf{Rate} \\
\midrule
Science & 20044 & 31.63\% \\
Math & 8789 & 13.87\% \\
SuperGPQA & 8409 & 13.27\% \\
MMLU\_pro & 6740 & 10.64\% \\
Korbench & 5877 & 9.27\% \\
BBH & 5745 & 9.07\% \\
Drop & 2974 & 4.69\% \\
SimpleQA & 1757 & 2.77\% \\
ChineseSimpleQA & 1215 & 1.92\% \\
GAOKAOBench & 849 & 1.34\% \\
Ceval & 484 & 0.76\% \\
Cmmlu & 308 & 0.49\% \\
ARC & 145 & 0.23\% \\
GPQA & 38 & 0.06\% \\
\midrule
\textbf{Total} & \textbf{63,374} & \textbf{100.00\%} \\
\bottomrule
\end{tabular}
\vspace{0.2cm}
\caption{Disagreement data distribution}
\label{tab:benchmark_allocation_new}
\end{table}

\begin{table}[ht]
\centering
\begin{tabular}{lrr}
\toprule
\textbf{Model} & \textbf{Count} & \textbf{Rate} \\
\midrule
NBG-Family & 13308 & 21.00\% \\
DeekSeek-Qwen-7B & 6337 & 10.00\% \\
GPT-OSS-20B & 5300 & 8.36\% \\
DeepSeek-Qwen-1.5B & 4874 & 7.69\% \\
Qwen3-30B-A3B & 4488 & 7.08\% \\
Llama3.1-8B & 4056 & 6.40\% \\
MiMo-7b & 3581 & 5.65\% \\
Qwen3-4B & 3405 & 5.37\% \\
Qwen2.5-7B & 2931 & 4.62\% \\
Qwen3-32B & 2917 & 4.60\% \\
Qwen3-1.7B & 2763 & 4.36\% \\
Gemma2-9B & 2474 & 3.90\% \\
Gemma2-2B & 2415 & 3.81\% \\
internLM-7B & 2277 & 3.59\% \\
QWEN3-8B & 2248 & 3.55\% \\
\midrule
\textbf{Total} & \textbf{63,374} & \textbf{100.00\%} \\
\bottomrule
\end{tabular}
\vspace{0.2cm}
\caption{Model responses count on disagreement data distribution (NBG-Family aggregated)}
\label{tab:model_ans_distribution_adjusted}
\end{table}

\begin{table}[ht]
\centering
\label{tab:cold-start-distribution} 
\begin{tabular}{lrr}
\toprule
\textbf{Source} & \textbf{Count} & \textbf{Rate} \\
\midrule
Python interpreter data& 2500 & 29.33\% \\
Unit conversion data & 1500 & 17.60\% \\
No Tool use data & 3000 & 35.20\% \\
Long-tail augment data & 1523 & 17.87\% \\
\midrule
\textbf{Total} & \textbf{8,523} & \textbf{100.00\%} \\
\bottomrule
\end{tabular}
\vspace{0.2cm}
\caption{Cold start data distribution}
\end{table}

\section{Training details}
\label{appendix:training details}
\subsection{Training \model-4B and \model-32B}
For labeling verifier \model-4B and \model-32B, we conducted supervised fine tuning on 73,714 samples, which contains 63.714 disagreement data, 2000 long-tail augmentation data and 8000 difficult-oriented augmentation data. These two verifiers are designed to output binary verification(\textit{Correct} and \textit{Incorrect}) without reasoning, maintaining superior efficiency while having performance comparable to \model-4B-Tool. Detailed training parameters are listed in Table \ref{tab:SFT}:

\begin{table}[htbp]
\centering
\begin{tabular}{ll}
\hline
\textbf{Parameter} & \textbf{Value} \\
\hline
BF16 & True \\
Learning Rate & $1 \times 10^{-5}$ \\
LR Scheduler Type & cosine\_with\_min\_lr \\
Maximum Sequence Length & 16384 \\
Training Epochs & 1 \\
Use Liger Kernel & True \\
Warmup Ratio & 0.01 \\
\hline
\end{tabular}
\vspace{0.2cm}
\caption{Cold start training configurations of \model-4B-Tool and supervised fine tuning configurations of \model-4B and \model-32B}
\label{tab:SFT}
\end{table}

\subsection{Ablation study of labeling verifier}
\begin{figure}[H]
        \centering
	\includegraphics[width=0.5\linewidth]{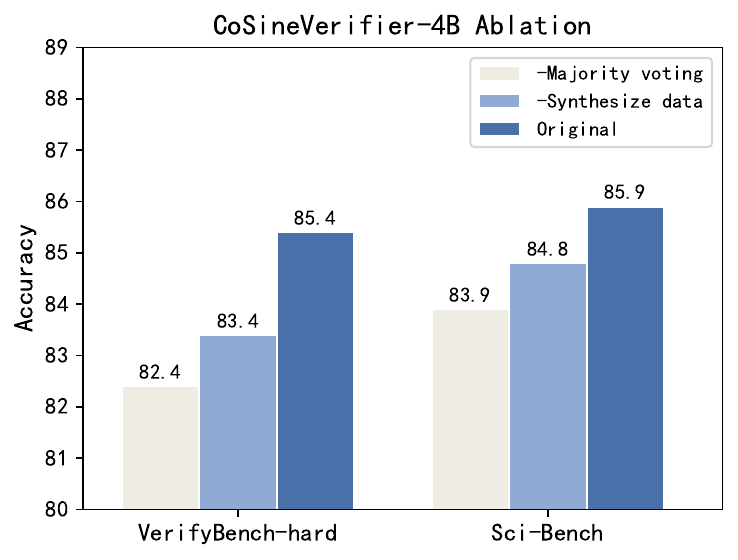}
        \caption{Ablation study of \model-4B}
 \label{fig:ablation_labeling}
\end{figure}

For \model-4B, we investigate the impact of removing: \textbf{(1) multi-turn majority voting}  and \textbf{(2) long-tail data synthesis}. As illustrated in Figure \ref{fig:ablation_labeling}, removing any model from \model consistently leads to performance degradation, demonstrating the necessity of our holistic framework design. Specifically, excluding multi-turn majority data results in the most significant performance deterioration across two major benchmarks. This finding strongly underscores the imperative of collecting reliable "golden" answers from diverse, real-world domains. In addition, removing long-tail synthesis data also causes a marked decline, showing the effectiveness of long-tail alignment for rare but high-value cases. In conclusion, only combing two components together yields the best performance

\subsection{Training details of \model-4B-Tool}
We detail the training configurations of \model-4B-Tool. For cold-start process, we use 8,523 samples for cold-start with long-tail augmentation data, and we then use 9,456 difficulty-oriented augmentation data for reinforcement learning. All training configurations are listed below. 

\begin{table}[H]
\centering
\begin{tabular}{ll}
\hline
\textbf{Parameters} & \textbf{Value} \\
\hline
BF16                  & True            \\
Train batch size    & 128            \\
Train epochs        & 10              \\
Learning rate       & 1e-6           \\
Max prompt length   & 16384           \\
Max response length & 2048          \\
$n$                   & 8              \\
$\epsilon_{\min}$     & 0.2            \\
$\epsilon_{\max}$     & 0.28           \\
Temperature           & 1.0            \\
\hline
\end{tabular}
\vspace{0.2cm}
\caption{RL training configurations of \model-4B-Tool}
\end{table}

\begin{table}[H]
\centering
\begin{tabular}{ll}
\hline
\textbf{Parameters} & \textbf{Value} \\
\hline
BF16                  & True            \\
Train batch size    & 128            \\
Train epochs        & 3              \\
Learning rate       & 1e-6           \\
Max prompt length   & 4096           \\
Max response length & 16384          \\
$n$                   & 8              \\
$\epsilon_{\min}$     & 0.2            \\
$\epsilon_{\max}$     & 0.28           \\
Temperature           & 1.0            \\
\hline
\end{tabular}
\vspace{0.2cm}
\caption{Training configurations of RLVR application}
\end{table}

\section{Case Study}
\label{appendix:case_study}
\newtcolorbox{promptbox}[1]{
  colback=gray!20,
  colframe=gray!75,
  boxrule=1pt,
  colbacktitle=blue!10!gray,   
  coltitle=white,               
  fonttitle=\bfseries,          
  title=#1,
  arc=3mm,
  fontupper=\ttfamily,
  boxsep=5pt,
  left=10pt,
  right=10pt,
  breakable, 
}

\begin{promptbox}{CosineVerifier-4B-Tool for calculation}
\textbf{Question:} A block of mass $m=100$ g and of negligible size slides down a wedge of mass $M=10$ kg, with angle above the horizontal $\theta=30^\circ$, and maximum height $h=2$ m. The block is free to move as well, and there is a coefficient of friction $\mu=\frac{1}{2}$ between the wedge and block. Initially, the block is at rest with respect to the wedge and sits at the top of the wedge. The wedge is moving at $v_0=1$ m/s with the point of the wedge moving "forward."

For this problem, use $g=10$ m/s$^2$.

Find the exact amount of time it takes for the block to slide all the way to the bottom of the wedge.

Remember to put your final answer within \\boxed{}

\par\noindent\rule{\textwidth}{0.4pt} 

\textbf{model\_answer:}

"model\_ans": "To solve this problem, we need to analyze the motion of the block and the wedge, taking into account the forces acting on them and the constraints of the system.

\#\#\# Step 1: Analyze the forces and motion

1. **Forces on the block:**
   - Gravity: \( mg \) acting downward.
   - Normal force: \( N \) perpendicular to the wedge's surface.
   - Friction force: \( f = \mu N \) acting up the wedge.

2. **Forces on the wedge:**
   - Gravity: \( Mg \) acting downward.
   - Normal force: \( N \) from the block.
   - Friction force: \( f = \mu N \) acting down the wedge.
   - Initial velocity: \( v_0 = 1 \) m/s.

\#\#\# Step 2: Set up the equations of motion

1. **For the block:**
   - Along the wedge: \( ma_{\text{block}} = mg \sin \theta - f \)
   - Perpendicular to the wedge: \( N = mg \cos \theta \)

2. **For the wedge:**
   - Horizontal direction: \( Ma_{\text{wedge}} = f \cos \theta - N \sin \theta \)

\#\#\# Step 3: Substitute the known values

1. **Normal force:**
   \[
   N = mg \cos \theta = 0.1 \times 10 \times \cos 30^\circ = 0.1 \times 10 \times \frac{\sqrt{3}}{2} = 0.5\sqrt{3} \text{ N}
   \]
2. **Friction force:**
   \[
   f = \mu N = \frac{1}{2} \times 0.5\sqrt{3} = 0.25\sqrt{3} \text{ N}
   \]
3. **Acceleration of the block along the wedge:**
   \[
   a_{\text{block}} = g \sin \theta - \frac{f}{m} = 10 \times \frac{1}{2} - \frac{0.25\sqrt{3}}{0.1} = 5 - 2.5\sqrt{3} \text{ m/s}^2
   \]

\#\#\# Step 4: Calculate the time to slide down

The block slides down a distance \( s = \frac{h}{\sin \theta} = \frac{2}{\sin 30^\circ} = 4 \) m along the wedge.

Using the kinematic equation:
\[
s = \frac{1}{2} a_{\text{block}} t^2
\]
\[
4 = \frac{1}{2} (5 - 2.5\sqrt{3}) t^2
\]
\[
t^2 = \frac{8}{5 - 2.5\sqrt{3}}
\]
\[
t = \sqrt{\frac{8}{5 - 2.5\sqrt{3}}}
\]

\#\#\# Step 5: Simplify the expression

To simplify, we can rationalize the denominator:
\[
t = \frac{8\sqrt{5 + 2.5\sqrt{3}}}{5}
\]

\#\#\# Final Answer:
\[
\boxed{\frac{8\sqrt{5 + 2.5\sqrt{3}}}{5}}
\]"

\par\noindent\rule{\textwidth}{0.4pt} 

\textbf{reference answer:}

t = $\sqrt{((1604(2+\sqrt{3})/505)}$ s

\par\noindent\rule{\textwidth}{0.4pt} 

\textbf{CosineVerifier-4B-Tool verification:}

<tool\_call> \par
\{"name": "python\_interpreter", "arguments": \{"code": "def main():\textbackslash n    import math\textbackslash n    cand = 8*math.sqrt(5 + 2.5*math.sqrt(3))/5\textbackslash n    gold = math.sqrt(1604*(2+math.sqrt(3))/505)\textbackslash n    print(cand, gold)\textbackslash n\textbackslash nif \_\_name\_\_ == \textbackslash "\_\_main\_\_\textbackslash ":\textbackslash n    main()"\}\}\par
</tool\_call>\par

<tool\_response>\par
\{"compile\_result": null, "execution\_info": \{"code\_length": 183, "execution\_phases": ["execution\_start", "execution\_end"], "execution\_start\_time": 1760942845.014331, "language": "python", "stdin\_length": 4, "stdin\_provided": true, "temp\_file": "/tmp/tmposhv6sxt.py", "total\_execution\_time": 0.028516054153442383, "warnings": []\}, "run\_result": \{"command": ["python3", "/tmp/tmposhv6sxt.py"], "execution\_time": 0.028277158737182617, "exit\_success": true, "return\_code": 0, "status": "Finished", "stderr": "", "stdin\_used": "None", "stdout": "4.887241058965765 3.4429464400122507\textbackslash n", "working\_directory": "/tmp"\}, "status": "Success"\}\par
</tool\_response>\par

\par\noindent\rule{\textwidth}{0.4pt} 

\textbf{assistant}

The candidate's final expression  
t = $8\sqrt{5 + 2.5\sqrt{3}}/5$  
is numerically $\approx$ 4.89 s.  

The reference expression  
t = $\sqrt{(1604(2+\sqrt{3})/505)}$  
is numerically $\approx$ 3.44 s.  

Since 4.89 $\neq$ 3.44, the candidate answer does not match the gold answer.  

\boxed{[Incorrect]}

\end{promptbox}

\begin{promptbox}{CosineVerifier-4B-Tool for string match}
\textbf{Question:} "Please generate the corresponding RNA sequence directly based on the dot-bracket notation of the secondary structure: .....((((.....)))).((((((.......((((((((((......))))))).)))
.......(((((((((((...)))))))))))............))))))"
\par\noindent\rule{\textwidth}{0.4pt} 

\textbf{model\_answer:} "AUUGCUCGAAUUUAUAGGACUUUUUUCUAUAAAGAAUAGUUUGGACUUGAAAUG
UAUUUAAAAACAAGAGGUUGGUAGAUUAUCAGCCUCUUUCUUGUCGUUGAAAAAG",

\par\noindent\rule{\textwidth}{0.4pt} 

\textbf{reference answer:}
ACGGGUUUCCCGGGAAACCCCAAAAUGGGGCCCCUGUCGGGUUUUAUUCCC
UGUCGUCGCCCUUUUUGGGA
\par\noindent\rule{\textwidth}{0.4pt} 

\textbf{CosineVerifier-4B-Tool verification:} 

<tool\_call>\par
\{"name": "python\_interpreter", "arguments": \{"code": "def main():\textbackslash n    correct = \textbackslash "AUUGCUCGAAUUUAUAGGACUUUUUUCUAUAAAGAAUAGUUUGG
ACUUGAAAUGUAUUUAAAAACAAGAGGUUGGUAGAUUAUCAGCCUCUUUCUUGUCGUUGAAAAAG\textbackslash "\textbackslash n    candidate = \textbackslash "ACGGGUUUCCCGGGAAACCCCAAAAUGGGGCCCCUGUCGGGUUUUAUUCCCUG
UCGUCGCCCUUUUUGGGA\textbackslash "\textbackslash print('match' if correct == candidate else 'no')\textbackslash n\textbackslash nif \_\_name\_\_ == \textbackslash "\_\_main\_\_\textbackslash ":\textbackslash n    main()"\}\}\par
</tool\_call>\par

<tool\_response>\par
\{"compile\_result": null, "execution\_info": \{"code\_length": 411, "execution\_phases": ["execution\_start", "execution\_end"], "execution\_start\_time": 1761468426.650931, "language": "python", "stdin\_length": 4, "stdin\_provided": true, "temp\_file": "/tmp/tmp4b5lvbkx.py", "total\_execution\_time": 0.028637170791625977, "warnings": []\}, "run\_result": \{"command": ["python3", "/tmp/tmp4b5lvbkx.py"], "execution\_time": 0.02838897705078125, "exit\_success": true, "return\_code": 0, "status": "Finished", "stderr": "", "stdin\_used": "None", "stdout": "False", "working\_directory": "/tmp"\}, "status": "Success"\}\par
</tool\_response>\par

\par\noindent\rule{\textwidth}{0.4pt} 

\textbf{assistant}

The candidate RNA sequence does not match the gold sequence at all; their nucleotide strings are completely different. \boxed{Incorrect}

\end{promptbox}

\begin{promptbox}{CosineVerifier-4B-Tool for unit conversion}
\textbf{Question:} "Please provide the LUMO energy value for this molecule.\textbackslash nO=C1C=C(CO)C(=N)N1"
\par\noindent\rule{\textwidth}{0.4pt} 

\textbf{model\_answer:} "-7.00\textasciitilde\textbackslash text\{kJ/mol\}"
\par\noindent\rule{\textwidth}{0.4pt} 

\textbf{reference answer:} "5.48e22 eV\textbackslash mole"
\par\noindent\rule{\textwidth}{0.4pt} 

\textbf{CosineVerifier-4B-Tool verification:}

<tool\_call>\par
\{"name": "unit\_conversion", "arguments": \{"value": 7.0, "source\_unit": "kJ/mol", "target\_unit": "eV/mole"\}\}\par
</tool\_call>\par

<tool\_response>\par
Unit parsed value: 4.369056352122534e+22 electron\_volt / mole\par
</tool\_response>\par

\par\noindent\rule{\textwidth}{0.4pt} 

\textbf{assistant}

The gold answer is 5.48e22 eV\textbackslash mole.
The candidate gives --7.00 kJ mol$^{-1}$.
The unit conversoin function tells us it's roughly equal to 4.369056352122534e+22 electron\_volt / mole .
This is nowhere near s 5.48e22 eV\textbackslash mole .

\boxed{[Incorrect]}

\end{promptbox}

\section{Prompts}
\label{appendix:prompt}

\begin{promptbox}{\model-4B-Tool verification prompt}

    You are a diligent and precise assistant tasked with evaluating the correctness of responses. You will receive a question, an output sentence, and the correct answer. Your task is to determine if the output sentence corectly answers the question based on the provided correct answer. You can perform a short tool call and a short reasoning process. After a short reasoing process, put your response in the final with either [Correct] or [Incorrect] wrapped in \boxed{}.
    
    \vspace{2mm}
    Evaluation Protocol:
    \begin{enumerate}
        \item Reference Standard:
        \begin{itemize}
            \item The standard (gold) answer is definitive and always correct.
            \item The question is always valid --- never challenge it.
            \item Allow equivalent meaning answers.
            \item Do not regenerate answers; only compare candidate's final answer with the gold answer.
            \item You only need to compare correct answer and output sentence, do not regenerate or judge correct answer.
        \end{itemize}
        
        \item Comparison Method:
        \begin{itemize}
            \item Analyze the question's requirements and the gold answer's structure.
            \item Determine if the question requires exact matching or allows equivalence.
            \item Compare ONLY the candidate's final answer. Ignore reasoning errors.
            \item Ignore differences in formatting or style.
            \item For math expressions: check algebraic equivalence step by step; if uncertain, test numerically at multiple points.
            \item For multiple-choice: only compare the final choice and its content.
        \end{itemize}
        
        \item Multi-part Answers:
        \begin{itemize}
            \item All parts must match the gold answer exactly.
            \item Partial matches are incorrect.
            \item If not specified, answer order may vary. For example, $\frac{{27}}{{7}}, -\frac{{8}}{{7}}$ and $-\frac{{8}}{{7}}, \frac{{27}}{{7}}$ are equivalent.
        \end{itemize}
    \end{enumerate}
    
    \vspace{2mm}
    Special considerations:
    \begin{enumerate}
        \item \textbf{Mathematical Problems}:
        \begin{itemize}
            \item If the formats differ but the answers are mathematically equivalent after simplfying or rounding to two decimal places (e.g. 2.909 vs $\frac{{32}}{{11}}$, $\frac{{32}}{{11}}$ vs $\frac{{96}}{{33}}$), respond with [Correct].
            \item You only need to verify the correctness of the mathematical expression, not values unrelated to the overall expression, such as the domain or units (e.g. 16 vs 16km, 20 vs 20db), these cases will be considered as [Correct].
            \item You may need to calculate the value or converse the value to different unit when needed to match the reference answer.
        \end{itemize}
        
        \item \textbf{Multi-choice questions}:
        \begin{itemize}
            \item If the question provides explicit candidate answers (e.g. multi-choice questions), the output will be considered correct if it clearly indicates the correct option's content or the correct option's code.
        \end{itemize}
        
        \item \textbf{Fact quuestions}:
        \begin{itemize}
            \item If the question provides fact-seeking answers, the output must align with the correct answer in content to be considered [Correct].
        \end{itemize}
        
        \item \textbf{Multiple Reference Answers}:
        \begin{itemize}
            \item If multiple reference answers are equivalent, just matching one answer will be considered [Correct].
            \item If multiple reference answers are inequivalent, only mathcing all answers will be considered [Correct].
        \end{itemize}
        
        \item \textbf{Ohter conditions}:
        \begin{itemize}
            \item If incomplete (cut off, unfinished sentence) $\rightarrow$ Label as [Incorrect].
            \item If repetitive (looping words/phrases) $\rightarrow$ Label as [Incorrect].
            \item Gives an answer but then negates it at the end. $\rightarrow$ Label as [Incorrect].
            \item Numerically correct but without units. $\rightarrow$ Label as [Correct].
        \end{itemize}
    \end{enumerate}
    
    \vspace{2mm}
    You can use following tools to help your verification process:
    \begin{enumerate}
        \item \textbf{Python Intepreter}: When you feel needed, you can use a python interpreter to help you determine your verification result.
        \item \textbf{Unit Conversion Tool}: When faced with different physical units, you can use a unit conversion tool to convert them into the same unit.
    \end{enumerate}
    
    \vspace{2mm}
    Question: """\{question\}"""  \par
    \vspace{2mm}
    Output sentence: """\{pred\}""" \par
    \vspace{2mm}
    Correct answer: \{reference\} \par
    \vspace{2mm}
    Judgement:
    
\end{promptbox}

\begin{promptbox}{Prompt for data augmentation}
Role:
You are an education expert.

\vspace{2mm}
Task:
Your student's assignment is to check whether a model's given answer to a question is consistent with the standard (reference) answer.
Your mission is to generate a new practice problem for the student based on their past mistakes, in order to test and strengthen their verification skills.

\vspace{2mm}
Input:
You will be given one examples of the student's incorrect exercises and its's mistake type.

\vspace{2mm}
Instructions:
First, analyze the given question, reference answer the model answer.
Second, Based on your analysis and the given mistake type, create a new practice question that specifically targets this weakness. Make sure the difficulty and style of this verification practice is on par with the mistake.

\vspace{2mm}
Output Format:
Please output the new practice problem in JSON format with the following fields:

\{\{
 "question": "...", \par
 "ref\_answer": "...", \par
 "model\_answer": "...", \par
 "ref\_judge": "Only in [Correct] or [Incorrect], represents the golden judge tag" \par
\}\}

\vspace{2mm}
Input Examples:
Mistake types: \{mistake\_type\}
<example1> \par
\{example1\} \par
</example1>\par

\vspace{2mm}
Your output:
\end{promptbox}

\begin{promptbox}{Error Analysis Prompt}
\# ROLE: You are an expert AI assistant specializing in error analysis.

\vspace{2mm}
\# TASK:
Your goal is to analyze an "Incorrect Verification" and categorize the reasoning mistake it contains. Compare the "Incorrect Verification" to the "Correct Verification" to identify the error types described below.

\vspace{2mm}
\# Instructions:
You should only choose the Error Categories below and donot create a new one. Focus only on the Category name, not the description in parentheses not the labeling numbers. 
Wrap your answer in the final with \\boxed\{\{\}\}

\vspace{2mm}
\# INPUTS: \par
\* **Question**: The original question that was asked. \par
\* **Model Answer**: The answer provided by a model, which is being evaluated. \par
\* **Reference Answer**: The ground-truth correct answer. \par
\* **Correct Verification (golden\_verify)**: The correct reasoning used to determine if the Model Answer is right or wrong. Only in '[Correct]' or '[Incorrect]'. \par
\* **Incorrect Verification (error\_verify)**: An flawed line of reasoning that you must analyze. Only in '[Correct]' or '[Incorrect]'.

\vspace{2mm}
---
\#\# Error Categories\#\#

1. Calculation Inaccuracy (The final numerical value is mathematically incorrect.) \par
2. Format Compliance Error (The answer's structure violates the required output format.)\par
3. Constraint Violation (The answer ignores a specific rule, e.g., "use each number once".)\par
4. Incorrect Simplification (Fails to simplify to minimal form or simplifies erroneously.)\par
5. Precision and Boundary Error (Incorrect rounding or definitions of interval endpoints, e.g., open/closed.)\par
6. Exact Match Failure (The response (e.g., a sequence) does not perfectly match the reference.)\par
7. Answer Completeness Error (Fails to provide all required answers, e.g., missing candidates or sub-parts.)\par
8. Missing Final Result (Provides only reasoning or code without the conclusive answer.)\par
9. Extraneous Content Error (The response includes correct data mixed with unrelated, incorrect information.)\par
10. Self-Correction Failure (The model initially answers correctly but changes to an incorrect answer.)\par
11. Refusal or Inconclusive Response (Fails to provide a clear answer or states it is unanswerable.)\par
12. Garbled or Corrupted Output (The response contains unreadable characters or significant noise.)\par
13. Truncated Response (The response is clearly cut off and incomplete.)\par
14. Meaningless Repetition Error (The response repeats text extensively, hiding any valid answer.)\par
15. Invalid Task or Reference (The error originates from a defective question or reference answer.)\par

\vspace{2mm}
**Question**: \{question\} \par
**Model Answer**: \{answer\} \par
**Reference Answer**: \{ref\_answer\} \par
**Correct Verification**: \{golden\_verify\} \par
**Incorrect Verification**: \{error\_verify\}\par

\vspace{2mm}
**Your classified error category:**
\end{promptbox}

\begin{promptbox}{All prompt prefixes for training data}
    \begin{enumerate}
        \item \texttt{"Let's think step by step and output the final answer within \textbackslash{}\textbackslash{}boxed\{\}.\textbackslash{}n"}

        \item \texttt{"Solve the following problem step by step. The last line of your response should be of the form Answer: \$Answer (without quotes) where \$Answer is the answer to the problem.\textbackslash{}n\textbackslash{}n"}

        \item \texttt{"Let's think step by step. \textbackslash{}n"}

        \item \texttt{"Please answer the following question. \textbackslash{}n"}

        \item \texttt{""}

        \item \texttt{"Answer the following multiple choice question. The last line of your response should be of the following format: 'Answer: \$LETTER' (without quotes) where LETTER is one of ABCD. Think step by step before answering.\textbackslash{}n\textbackslash{}n"}

        \item \texttt{"Answer the following multiple choice question. The last line of your response should be of the following format: 'Answer: \$LETTER' (without quotes) where LETTER is one of ABCD.\textbackslash{}n\textbackslash{}n"}
    \end{enumerate}
\end{promptbox}

\end{document}